\documentclass{article}

\usepackage{microtype}
\usepackage{graphicx}
\usepackage{subfigure}
\usepackage{booktabs} 

\usepackage{hyperref}

\usepackage[accepted]{icml2021}

\usepackage{amsmath,amssymb}
\usepackage{mathtools}

\allowdisplaybreaks

\icmltitlerunning{Reinforced Imitation Learning by Free Energy Principle}

\begin{document}

\twocolumn[
\icmltitle{Reinforced Imitation Learning by Free Energy Principle}



\icmlsetsymbol{equal}{*}

\begin{icmlauthorlist}
\icmlauthor{Ryoya Ogishima}{ut}
\icmlauthor{Izumi Karino}{ut}
\icmlauthor{Yasuo Kuniyoshi}{ut}
\end{icmlauthorlist}

\icmlaffiliation{ut}{Graduate School of Information Science and Technology, The University of Tokyo, Tokyo, Japan}

\icmlcorrespondingauthor{Ryoya Ogishima}{ogishima@isi.imi.i.u-tokyo.ac.jp}

\icmlkeywords{Free Energy Principle, Imitation, Reinforcement Learning}

\vskip 0.3in
]



\printAffiliationsAndNotice{}  

\begin{abstract}
Reinforcement Learning (RL) requires a large amount of exploration especially in sparse-reward settings. Imitation Learning (IL) can learn from expert demonstrations without exploration, but it never exceeds the expert's performance and is also vulnerable to distributional shift between demonstration and execution. In this paper, we radically unify RL and IL based on Free Energy Principle (FEP). FEP is a unified Bayesian theory of the brain that explains perception, action and model learning by a common fundamental principle. We present a theoretical extension of FEP and derive an algorithm in which an agent learns the world model that internalizes expert demonstrations and at the same time uses the model to infer the current and future states and actions that maximize rewards. The algorithm thus reduces exploration costs by partially imitating experts as well as maximizing its return in a seamless way, resulting in a higher performance than the suboptimal expert. Our experimental results show that this approach is promising in visual control tasks especially in sparse-reward environments.
\end{abstract}

\section{Introduction}

Reinforcement Learning (RL) autonomously explores to maximize rewards, even achieving super-human performances in certain tasks\cite{sutton1998introduction, silver2016mastering}. It can also transfer the acquired policy to new tasks/environments with additional explorations.
However, in realistic tasks, RL often requires an excess amount of explorations especially in sparse-reward settings, and even when it succeeds in reward maximization, the acquired policy sometimes severely deviates from the intention of the reward designer.

Imitation Learning (IL) learns a policy to mimic the trajectories demonstrated by an expert\cite{pomerleau1991efficient}. Therefore it does not require explorations nor a careful design of a reward function. However, the policy acquired by IL never exhibits the performance exceeding that of the suboptimal expert, and it is also vulnerable to realistic setting with distributional shift between the demonstration and execution environments or perturbation and noise.

Since the pros and cons of RL and IL are mutually compensating, a natural consequence would be to combine the two methods. Several work have been reported along this idea \cite{verma2019imitation,pfeiffer2018reinforced,sun2018dual,rhinehart2018deep}.
However, there still remains an important problem. That is, a truly seamless unification of IL and RL on a common theoretical ground to make the best out of mutual leverages, and dealing with another realistic setting which is partial observability as in POMDP (Partially Observable Markov Decision Process), particularly with high-dimensional image inputs. It is widely assumed that introducing a generative model of the world in terms of latent variables is a promising approach to POMDP.

Recent work in model-based RL succeeds in latent planning from
high-dimensional image inputs by incorporating latent dynamics models. Behaviors
can be derived either by imagined-reward maximization \cite{ha2018world, hafner2019dream} or by online planning \cite{hafner2018learning}.
Although solving high dimensional visual control tasks with model-based methods is becoming feasible, prior methods have not been combined with IL.

Free Energy Principle (FEP), a unified brain theory in computational neuroscience
 explains perception, action and model learning in a Bayesian probabilistic way \cite{friston2006free, friston2010free}.
In FEP, the brain has a generative model of the world and computes
a mathematical amount called Free Energy using the model prediction and sensory inputs to the brain.
By minimizing the Free Energy, the brain achieves model learning and behavior learning. Therefore it has a potential to fundamentally unify IL and RL on the common theoretical ground.
However, prior work on FEP dealt with limited situations where a part of
the generative model is given and the task is very low dimensional.
As there are a lot in common between FEP and variational inference in machine learning, recent advancements in
deep learning and latent variable models could be applied to scale up FEP agents to be compatible with high dimensional tasks.

In this paper, we propose Deep Free Energy Network (FENet), an agent that combines the advantages of IL and RL so that the initial policy is learned from suboptimal expert data without the need of exploration or detailed reward crafting, then it is further improved from sparsely specified reward functions to exceed the suboptimal expert performance.

The key contributions of this work are summarized as follows:
\begin{itemize}
  \item \textbf{Extension of Free Energy Principle:} \\
  We theoretically extend Free Energy Principle, introducing policy prior and policy
  posterior to combine IL and RL. We implement the proposed method on top of
  Recurrent State Space Model \cite{hafner2018learning}, a latent dynamics model with both deterministic and stochastic components.
  \item \textbf{Visual control tasks in realistic problem settings:} \\
  We solve Cheetah-run, Walker-walk, and Quadruped-walk tasks from DeepMind Control Suite \cite{tassa2018deepmind}.
  We do not only use the default problem settings, but we also set up problems with sparse rewards and with suboptimal experts. We demonstrate that our agent outperforms model-based RL using Recurrent State Space Model in sparse-reward settings. We also show that our agent can achieve higher returns than Behavioral Cloning (IL) with suboptimal experts.
\end{itemize}

\section{Backgrounds on Free Energy Principle}
\subsection{Problem setups}
We formulate visual control as a partially observable Markov decision process (POMDP)
with discrete time steps $t$, observations $o_t$, hidden states $s_t$, continuous
action vectors $a_t$, and scalar rewards $r_t$. The goal is to develop an agent
that maximizes expected return $\mathbb{E}[\sum_{t=1}^T r_t]$.

\subsection{Free Energy Principle}
\label{fep}
Perception, action and model learning are all achieved by minimizing the same objective function, Free Energy \cite{friston2006free, friston2010free}.
In FEP, the agent is equipped with a generative model of the world, using a prior
$p(s_t)$ and a likelihood $p(o_t|s_t)$.
\begin{align}
  p(o_t,s_t) = p(o_t|s_t)p(s_t)
\end{align}
\textbf{Perceptual Inference} \quad
Under the generative model, the posterior probability of hidden states given observations is
calculated with Bayes' theorem as follows.
\begin{align}
  p(s_t|o_t) = \frac{p(o_t|s_t)p(s_t)}{p(o_t)}, \quad
  p(o_t) = \int p(o_t|s_t)p(s_t) ds
\end{align}
Since we cannot compute $p(o_t)$ due to the integral, we think of approximating $p(s_t|o_t)$ with
a variational posterior $q(s_t)$ by minimizing KL divergence $KL(q(s_t)||p(s_t|o_t))$.
\begin{align}
  KL(q(s_t)||p(s_t|o_t)) &= \ln p(o_t) + KL(q(s_t)||p(o_t,s_t)) \label{posterior}\\
  F_t &= KL(q(s_t)||p(o_t,s_t)) \label{fe}
\end{align}
The Free Energy is defined as (eq.\ref{fe}). Since $p(o_t)$ does not depend on $s_t$, we can minimize (eq.\ref{posterior})
w.r.t. the parameters of the variational posterior by minimizing the Free Energy.
Thus, the agent can infer the hidden states of the observations by minimizing $F_t$. This process is called 'perceptual inference' in FEP.

\textbf{Perceptual Learning} \quad
Free Energy is the same amount as negative Evidence Lower Bound (ELBO) in variational inference
often seen in machine learning as follows.
\begin{align}
  \ln p(o_t) \geq -F_t
\end{align}
By minimizing $F_t$ w.r.t. the parameters of the prior and the likelihood, the generative
model learns to best explain the observations. This process is called 'perceptual learning' in FEP.

\textbf{Active Inference} \quad
We can assume that the prior is conditioned on the hidden states and actions at the previous time step as follows.
\begin{align}
p(s_t) \coloneqq p(s_t|s_{t-1}, a_{t-1})
\end{align}
The agent can change the future by choosing actions. Suppose the agent chooses $a_t$ when it is at $s_t$,
the prior can predict the next hidden state $s_{t+1}$. Thus, we can think of the Expected
Free Energy $G_{t+1}$ at the next time step $t+1$ as follows \cite{friston2015active}.
\begin{align}
  G_{t+1} &= \mathbb{E}_{p(o_{t+1}|s_{t+1})}[KL(q(s_{t+1})||p(o_{t+1}, s_{t+1}))] \nonumber \\
    &= \mathbb{E}_{q(s_{t+1})p(o_{t+1}|s_{t+1})}[\ln q(s_{t+1}) - \ln p(o_{t+1}, s_{t+1})] \label{obs_expectation} \\
    &= \mathbb{E}_{q(s_{t+1})p(o_{t+1}|s_{t+1})}[\ln q(s_{t+1}) - \ln p(s_{t+1}|o_{t+1}) \nonumber \\
    &\quad - \ln p(o_{t+1})] \nonumber \\
    &\approx \mathbb{E}_{q(o_{t+1}, s_{t+1})}[\ln q(s_{t+1}) - \ln q(s_{t+1}|o_{t+1}) \nonumber \\
    &\quad - \ln p(o_{t+1})] \label{expectedFE} \\
    &= \mathbb{E}_{q(o_{t+1})}[-KL(q(s_{t+1}|o_{t+1})||q(s_{t+1})) \nonumber \\
    &\quad - \ln p(o_{t+1})] \label{expectedFE2}
\end{align}
Since the agent has not experienced time step ${t+1}$ yet and has not received observations $o_{t+1}$,
we take expectation over $o_{t+1}$ using the likelihood $p(o_{t+1}|s_{t+1})$ as (eq.\ref{obs_expectation}). In (eq.\ref{expectedFE}),
we define the likelihood $q(o_{t+1}|s_{t+1}) = p(o_{t+1}|s_{t+1})$ and approximate the posterior $p(s_{t+1}|o_{t+1})$ as the variational posterior
$q(s_{t+1}|o_{t+1})$. According to the complete class theorem \cite{friston2012active},
any scalar rewards can be encoded as observation priors using $p(o) \propto \exp r(o)$ and
the second term in (eq.\ref{expectedFE2}) becomes a goal-directed value. This observation prior $p(o_{t+1})$ can also be regarded as the probability of optimality variable $p(\mathcal{O}_{t+1}=1|o_{t+1})$, where the binary optimality variable $\mathcal{O}_{t+1}=1$ denotes that time step $t+1$ is optimal and $\mathcal{O}_{t+1}=0$ denotes that it is not optimal as introduced in the context of control as probabilistic inference\cite{levine2018reinforcement}. The first term
in (eq.\ref{expectedFE2}) is called epistemic value that works as intrinsic motivation to further explore the world.
Minimization of $-KL(q(s_{t+1}|o_{t+1})||q(s_{t+1}))$ means that the agent tries to experience as different states $s_{t+1}$ as possible given some imagined observations $o_{t+1}$.
By minimizing the Expected Free Energy, the agent can infer the actions that explores
the world and maximize rewards. This process is called 'active inference'.

\section{Deep Free Energy Network (FENet)}
Perceptual learning deals with learning the generative model to best explain the agent's sensory inputs.
If we think of not only observations but also actions demonstrated by the expert as a part of the sensory inputs,
we can explain IL by using the concept of perceptual learning. In other words, this is a process of learning the world model that internalizes expert demonstrations as passive dynamics. Active inference deals with
exploration and reward maximization, so it is compatible with reinforcement learning.
By minimizing the same objective function, the Free Energy, we can deal with both IL and RL.

In this section, we first extend the Free Energy so that actions are a part of the sensory inputs to accommodate both IL and RL. For this purpose, we introduce a policy prior for IL and a policy posterior for RL.
Second, we use the extended Free Energy to derive and extend the Expected Free Energy in two ways. One is for calculation with given expert data for IL, and the other is for calculation with collected agent data for RL. Finally, we explain a detailed network design to implement the proposed method for solving image control tasks.

\subsection{Introducing a policy prior and a policy posterior}
\textbf{Free Energy} \quad
We extend the Free Energy from (eq.\ref{fe}) so that actions are a part of the sensory
inputs that the generative model tries to explain.
\begin{align}
F_t &= KL(q(s_t)||p(o_t,s_t,a_t)) \\
  &= KL(q(s_t)||p(o_t|s_t)p(a_t|s_t)p(s_t|s_{t-1},a_{t-1})) \\
  &= \mathbb{E}_{q(s_t)}[\ln \frac{q(s_t)}{p(o_t|s_t)p(a_t|s_t)p(s_t|s_{t-1},a_{t-1})}] \\
  &= \mathbb{E}_{q(s_t)}[-\ln p(o_t|s_t) - \ln p(a_t|s_t) + \ln q(s_t) \nonumber \\
  &\quad - \ln p(s_t|s_{t-1},a_{t-1})] \label{proposed_fe} \\
  &= \mathbb{E}_{q(s_t)}[-\ln p(o_t|s_t) - \ln p(a_t|s_t)] \nonumber \\
  &\quad + KL(q(s_t)||p(s_t|s_{t-1},a_{t-1}))
\end{align}
We define $p(a_t|s_t)$ as a policy prior. When the agent observes expert trajectories,
by minimizing $F_t$ w.r.t. the policy prior parameters, the policy prior will be learned so that it can best explain
the experts. By minizing $F_t$ w.r.t. the parameters of the state prior $p(s_t|s_{t-1},a_{t-1})$ and the observation likelihood $p(o_t|s_t)$, the world model is learned as explained as perceptual learning in Section~\ref{fep}. Besides the policy prior, we introduce and define a policy posterior $q(a_t|s_t)$, which
is the very policy that the agent samples actions from when interacting with its environments and that the agent uses to imagine the future observations and rewards. We explain how to learn
the policy posterior in the following.

\textbf{Expected Free Energy for Imitation Learning (IL)} \quad
In a similar manner to active inference in Section~\ref{fep}, we think of the Expected Free Energy $G_{t+1}$
at the next time step $t+1$, but this time we take expectation
over the policy posterior $q(a_t|s_t)$ because $G_{t+1}$ is a value expected under the next actions.
Note that in Section~\ref{fep} $a_t$ was given as a certain value input, but here $a_t$ is sampled from the policy posterior.
We calculate the expected state at time step ${t+1}$ as follows.
\begin{align}
  q(s_{t+1}) &= \mathbb{E}_{q(s_t)q(a_t|s_t)}[p(s_{t+1}|s_t,a_t)] \\
  q(o_{t+1}, s_{t+1}, a_{t+1}) &= p(o_{t+1}|s_{t+1})q(a_{t+1}|s_{t+1})q(s_{t+1})
\end{align}
We derive the Expected Free Energy from (eq.\ref{proposed_fe}) as follows.
\begin{align}
  G_{t+1}^{IL} &= \mathbb{E}_{q(o_{t+1}, s_{t+1}, a_{t+1})}[- \ln p(o_{t+1}|s_{t+1}) \nonumber \\
  &\quad - \ln p(a_{t+1}|s_{t+1}) + \ln q(s_{t+1}) - \ln p(s_{t+1}|s_t,a_t)] \label{expectedIL}\\
  &= \mathbb{E}_{q(o_{t+1}, s_{t+1}, a_{t+1})}[- \ln p(o_{t+1}|s_{t+1}) \nonumber \\
  &\quad - \ln p(a_{t+1}|s_{t+1}) + 0] \label{expectedIL2}\\
  &= \mathbb{E}_{q(o_{t+1}, s_{t+1})}[-\ln p(o_{t+1}|s_{t+1}) \nonumber \\
  &\quad + \mathbb{E}_{q(a_{t+1}|s_{t+1})}[-\ln p(a_{t+1}|s_{t+1})]] \\
  &= \mathbb{E}_{q(s_{t+1})}[\mathcal{H}[p(o_{t+1}|s_{t+1})] \nonumber \\
  &\quad + \mathbb{E}_{q(a_{t+1}|s_{t+1})}[-\ln p(a_{t+1}|s_{t+1})]] \label{imitationG}
\end{align}
In (eq.\ref{expectedIL2}), $q(s_{t+1})$ and $p(s_{t+1}|s_t,a_t)$ are both state prior prediction at future time step $t+1$, and they are regarded as the same value. In (eq.\ref{imitationG}), the first term is the entropy of the observation likelihood,
and the second term is the negative likelihood of the policy prior expected under the policy posterior.
By minimizing $G_{t+1}^{IL}$, the agent learns the policy posterior so that it matches
the policy prior which has been learned through minimizing $F_t$ to encode the experts' behavior.

\textbf{Expected Free Energy for RL} \quad
We can get the Expected Free Energy in a different way that has a reward
component $r(o_{t+1})$ leading to the policy posterior maximizing rewards.
We derive the Expected Free Energy from (eq.\ref{expectedIL}) as follows.
\begin{align}
  G_{t+1}^{RL} &= \mathbb{E}_{q(o_{t+1}, s_{t+1}, a_{t+1})}[-\ln p(o_{t+1}, s_{t+1}) \nonumber \\
         &\quad - \ln p(a_{t+1}|s_{t+1}) + \ln q(s_{t+1})] \\
         &= \mathbb{E}_{q(o_{t+1}, s_{t+1}, a_{t+1})}[-\ln p(s_{t+1}|o_{t+1})) \nonumber \\
         &\quad -\ln p(a_{t+1}|s_{t+1}) + \ln q(s_{t+1}) - \ln p(o_{t+1})] \\
         &\approx \mathbb{E}_{q(o_{t+1}, s_{t+1}, a_{t+1})}[-\ln q(s_{t+1}|o_{t+1})) \nonumber \\
         &\quad -\ln p(a_{t+1}|s_{t+1}) + \ln q(s_{t+1}) - \ln p(o_{t+1})] \label{expectedRL}\\
         &= \mathbb{E}_{q(o_{t+1}, s_{t+1})}[- \ln q(s_{t+1}|o_{t+1}) - \ln p(o_{t+1}) \nonumber \\
         &\quad + \mathbb{E}_{q(a_{t+1}|s_{t+1})}[-\ln p(a_{t+1}|s_{t+1})] + \ln q(s_{t+1})] \\
         &= \mathbb{E}_{q(o_{t+1})}[-KL(q(s_{t+1}|o_{t+1})||q(s_{t+1})) - \ln p(o_{t+1})] \nonumber \\
         &\quad + \mathbb{E}_{q(s_{t+1})}[\mathbb{E}_{q(a_{t+1}|s_{t+1})}[-\ln p(a_{t+1}|s_{t+1})]] \\
         &\approx \mathbb{E}_{q(o_{t+1})}[-KL(q(s_{t+1}|o_{t+1})||q(s_{t+1})) - r(o_{t+1})] \nonumber \\
         &\quad + \mathbb{E}_{q(s_{t+1})}[\mathbb{E}_{q(a_{t+1}|s_{t+1})}[-\ln p(a_{t+1}|s_{t+1})]] \label{rlG}
\end{align}
In (eq.\ref{expectedRL}), we approximate $p(s_{t+1}|o_{t+1})$ as $q(s_{t+1}|o_{t+1})$ similarly to (eq.\ref{expectedFE}). This is to approximate true state posterior as variational state posterior. In (eq.\ref{rlG}), as is explained in active inference in Section~\ref{fep}, we use $p(o) \propto \exp r(o)$ for the approximation. The first KL term is the epistemic value that lets the agent explore the world,
the second term is the expected reward under the action sampled from the policy posterior,
and the last term is the likelihood of the policy prior expected under the policy posterior. Note that $q(o_{t+1})$ in (eq.\ref{rlG}) can be calculated as follows.
\begin{align}
  q(o_{t+1}) = \mathbb{E}_{q(s_{t+1})}[p(o_{t+1}|s_{t+1})]
\end{align}
By minimizing $G_{t+1}^{RL}$, the agent learns the policy posterior so that it
explores the world and maximizes the reward as long as it does not deviate
too much from the policy prior which has encoded experts' behavior through minimizing $F_t$.

In summary, $G_{t+1}^{IL}$ and $G_{t+1}^{RL}$ are both derived from the same Free Energy $F_t$, but they are different kinds of derivations to accommodate the data inputs required for IL and RL respectively.

\begin{figure}[t]
\begin{center}
\includegraphics[width=\columnwidth]{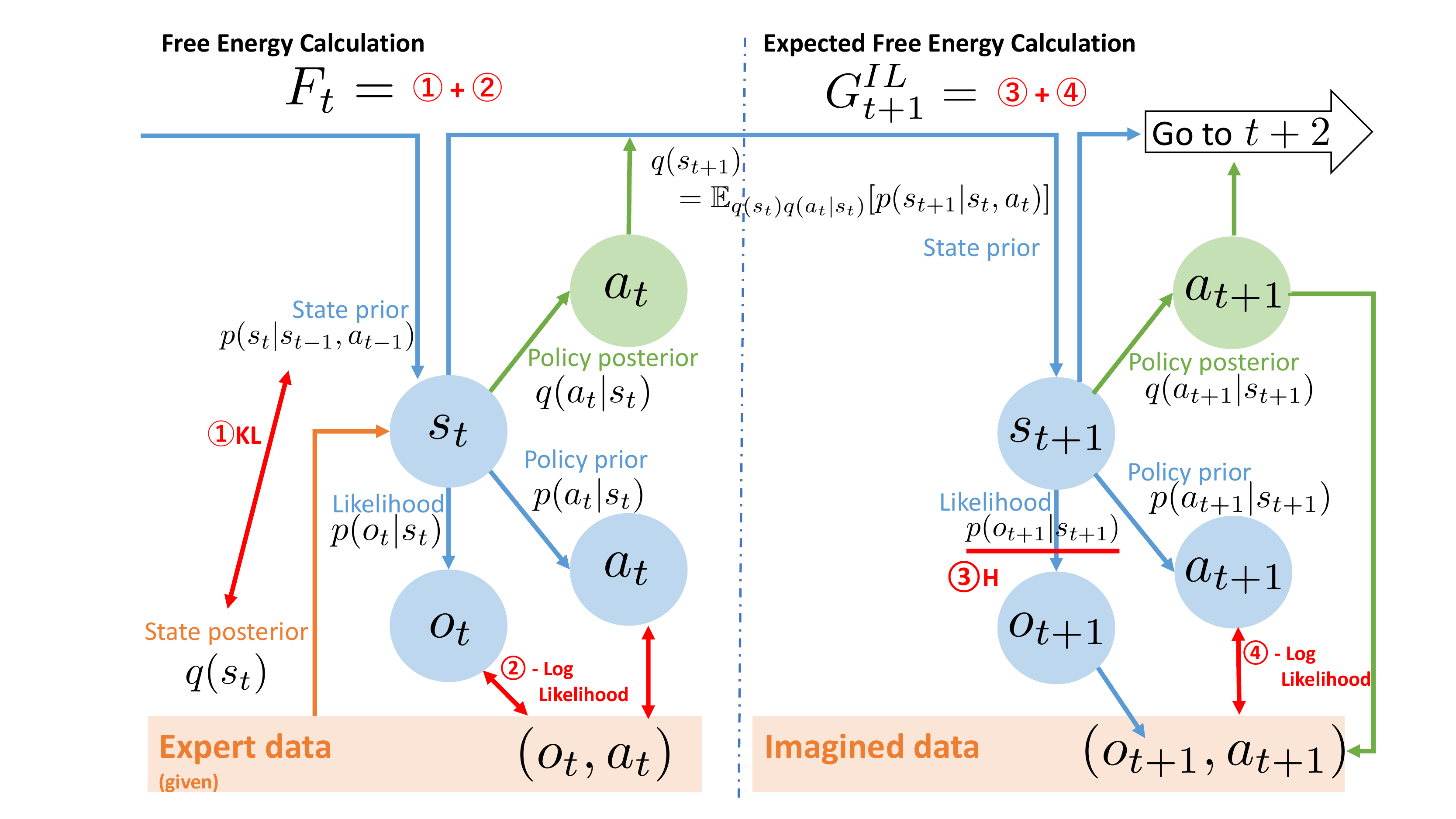}
\end{center}
\caption{Deep Free Energy Network (FENet) calculation process for IL phase.}
\label{model_il}
\end{figure}

\begin{figure}[t]
\begin{center}
\includegraphics[width=\columnwidth]{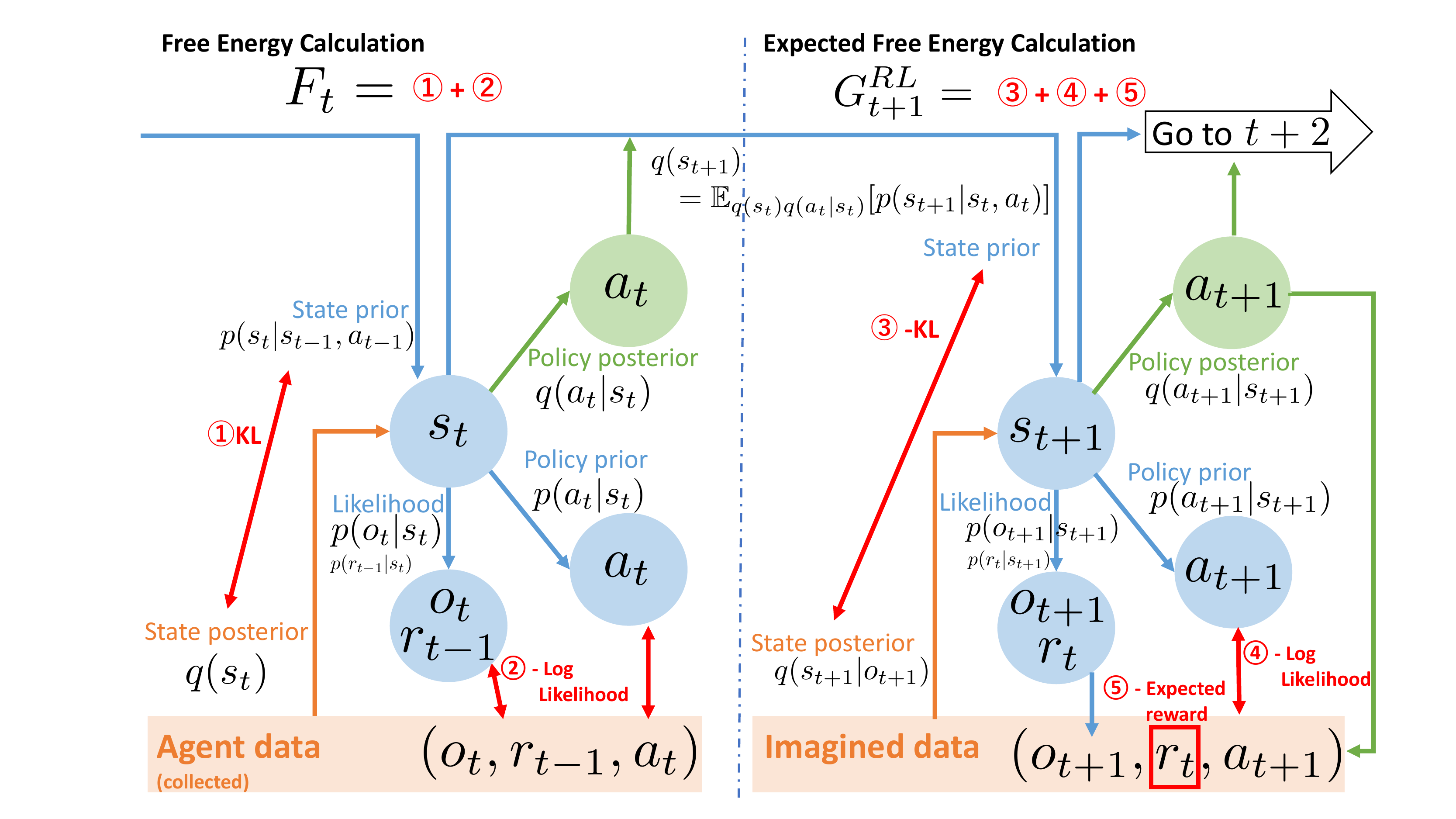}
\end{center}
\caption{Deep Free Energy Network (FENet) calculation process for RL phase.}
\label{model_rl}
\end{figure}

\subsection{IL and RL objectives}
To realize IL and RL at the same time, we propose that the agent calculate Free Energy-based losses for given expert data and collected agent data. The overall loss function of Deep Free Energy Network should minimize is as follows.
\begin{align}
  \mathcal{F}_{IL} + \mathcal{F}_{RL}
\end{align}
when, for given expert data,
\begin{align}
  \mathcal{F}_{IL} = F_t + \sum_{\tau=t+1}^\infty \gamma^{\tau-t-1} G_\tau^{IL}
\end{align}
for collected agent data,
\begin{align}
  \mathcal{F}_{RL} = F_t + \sum_{\tau=t+1}^\infty \gamma^{\tau-t-1} G_\tau^{RL}
\end{align}
Note that the Expected Free Energy at $t+1$ to $\infty$ are calculated to account for the long term future and that $\gamma$ is a discount factor as in the case of general RL algorithms. The overall Free Energy calculation process is shown in Figure~\ref{model_il} and Figure~\ref{model_rl}.

\subsection{Network Design for Implementation}
\textbf{Value function} \quad
As it is impossible to sum over infinity time steps, we introduce an Expected Free
Energy Value function $V(s_{t+1})$ to estimate the cumulative Expected Free Energy.
Similarly to the case of Temporal Difference learning of Deep Q Network \cite{mnih2013playing}, we use a
target network $V_{targ}(s_{t+2})$ to stabilize the learning process and define
the loss for learning the value function as follows.
\begin{align}
    \mathcal{L} = ||G_{t+1} + \gamma V_{targ}(s_{t+2}) - V(s_{t+1})||^2
\end{align}
To reduce the number of parameters of the networks, we made a design choice that the agent uses the value function only for RL, and not for IL. In IL, we use only the value of the Expected Free Energy $G_{t+1}$ at the next time step $t+1$ and ignore the time steps from $t+2$ to infinity. Note that we use a notation that the origin of time step $t$ is set at the time of the expert data the agent learns from in every iteration of learning. This means that the time steps from $t+1$ to infinity are all imagined time because the agent has not observed the time yet. Therefore, IL from $F_t + G^{IL}_{t+1}$ without $t+2$ to infinity still handles all time series data of expert demonstrations. It just does not predict more than 1 time step ahead. While IL can be achieved without long term prediction, RL needs the value function to
predict rewards in the long-term future to avoid a local minimum behavior and achieve the desired goal.

\textbf{Recurrent State Space Model} \quad
We made a design choice for the network implementation to use Recurrent State Space Model
\cite{hafner2018learning}, a latent dynamics model with both deterministic
and stochastic components. In this model, the hidden states $s_t$ are split into
two parts: stochastic hidden states $u_t$ and deterministic hidden
states $h_t$. The deterministic transition of $h_t$ is modeled using Recurrent Neural Networks (RNN) $f$ as follows.
\begin{align}
  h_t = f(h_{t-1}, u_{t-1}, a_{t-1})
\end{align}

We model the probabilities in Deep Free Energy Networks as follows.
\begin{align}
  &\text{State prior} &\quad& p_\theta(u_t|h_t) \\
  &\text{Observation likelihood} &\quad& p_\theta(o_t|u_t, h_t) \\
  &\text{Reward likelihood} &\quad& p_\theta(r_{t-1}|u_t, h_t) \\
  &\text{State posterior} &\quad& q_\phi(u_t|h_t, o_t) \\
  &\text{Policy prior} &\quad& p_\theta(a_t|u_t, h_t) \\
  &\text{Policy posterior} &\quad& q_\psi(a_t|u_t, h_t) \\
  &\text{Value network} &\quad& V_\omega(u_t) \\
  &\text{Target Value Network} &\quad& V_{\omega_{targ}}(u_t)
\end{align}
We model these probabilities as feedforward Neural Networks that output the mean
and standard deviation of the random variables according to the Gaussian distribution. For example, the policy posterior is modeled as a network that takes $u_t$ and $h_t$ as inputs and calculates through several hidden layers and outputs the Gaussian distribution of $a_t$. Note that $\theta, \phi, \psi, \omega$ are a group or set of network parameters to be learned such as network weights. For example, $theta$ is a group or set of all parameters consisting the probabilities of state prior, observation/reward likelihood and policy prior. Using the network parameters, the objective loss functions can be written as follows.
\begin{align}
  \mathcal{F}_{IL} &= F_t + G_{t+1}^{IL} \label{fil} \\
  \mathcal{F}_{RL} &= F_t + G_{t+1}^{RL} + \gamma V_{\omega_{targ}}(u_{t+2}) \label{frl} \\
  \mathcal{L} &= ||G_{t+1}^{RL} + \gamma V_{\omega_{targ}}(u_{t+2}) - V_\omega(u_{t+1})||^2 \label{value_loss} \\
  \text{when} \nonumber \\
  F_t &= \mathbb{E}_{q_\phi(u_t|h_t, o_t)}[-\ln p_\theta(o_t|u_t, h_t) - \ln p_\theta(a_t|u_t, h_t)] \nonumber \\
  &\quad + KL(q_\phi(u_t|h_t, o_t)||p_\theta(u_t|h_t)) \\
  G_{t+1}^{IL} &= \mathbb{E}_{q_(u_{t+1})}[\mathcal{H}[p_\theta(o_{t+1}|u_{t+1}, h_{t+1})] \nonumber \\
  &\quad + KL(q_\psi(a_{t+1}|u_{t+1}, h_{t+1})||p_\theta(a_{t+1}|u_{t+1}, h_{t+1}))] \\
  G_{t+1}^{RL} &= \mathbb{E}_{q(o_{t+1})}[-KL(q_\phi(u_{t+1}|h_{t+1}, o_{t+1})||q(u_{t+1})) \nonumber \\
  &\quad - p_\theta(r_t|u_{t+1}, h_{t+1})] + \mathbb{E}_{q(u_{t+1})}[ \nonumber \\
  &\quad KL(q_\psi(a_{t+1}|u_{t+1}, h_{t+1})||p_\theta(a_{t+1}|u_{t+1}, h_{t+1}))] \nonumber \\
  &\quad + \gamma V_{\omega_{targ}}(u_{t+2}) \\
  q(u_{t+1}) &= \mathbb{E}_{q_\phi(u_t|h_t, o_t)q_\psi(a_t|u_t, h_t)}[p_\theta(u_{t+1}|h_{t+1})] \\
  q(o_{t+1}) &= \mathbb{E}_{q(u_{t+1})}[p_\theta(o_{t+1}|u_{t+1}, h_{t+1})]
\end{align}
Algorithm~\ref{alg:FENet} shows overall calculations using these losses. The agent minimizes $\mathcal{F}_{IL}$ for expert data $\mathcal{D}_E$ and the agent minimizes $\mathcal{F}_{RL}$ for agent data $\mathcal{D}_A$ that the agent collects on its own.

\section{Experiments}
We evaluate FENet on three continuous control tasks from images. We compare our model
with model-based RL and model-based RL with demonstrations in dense and sparse reward setting when optimal expert is available. Then we compare our model with IL methods when only suboptimal experts are available. Finally, we investigate the merits of combining IL and RL as an ablation study.

\begin{figure}[h]
  \centering
  \subfigure[Cheetah-run]{
    \includegraphics[width=0.3\columnwidth]{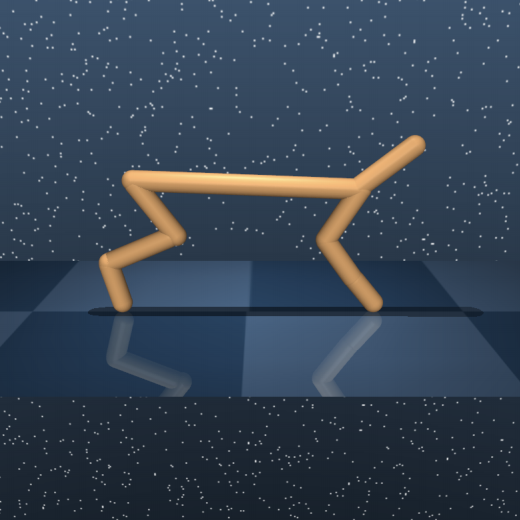}
  }
  \hfill
  \centering
  \subfigure[Walker-walk]{
    \includegraphics[width=0.3\columnwidth]{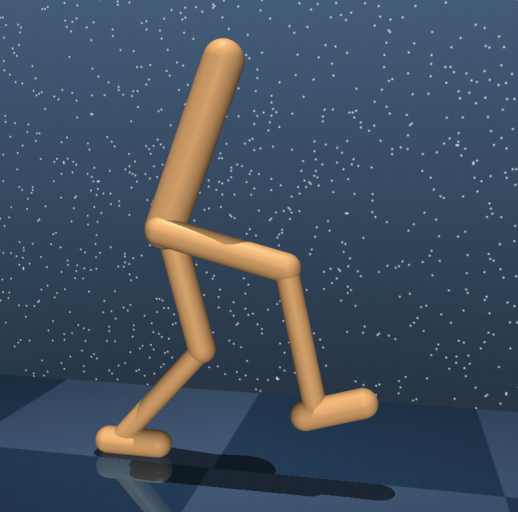}
  }
  \hfill
  \centering
  \subfigure[Quadruped-walk]{
    \includegraphics[width=0.3\columnwidth]{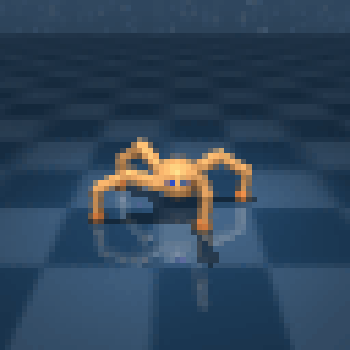}
  }
  \caption{Image-based control tasks used in our experiments.}
  \label{dm_control}
\end{figure}

\textbf{Control tasks} \quad
We used Cheetah-run, Walker-walk, and Quadruped-walk tasks, image-based continuous
control tasks of DeepMind Control Suite \cite{tassa2018deepmind} shown in Figure~\ref{dm_control}.
The agent gets rewards ranging from $0$ to $1$. Quadruped-walk is the most difficult as it has more action dimensions than the others.
Walker-walk is more challenging than Cheehtah-run because an agent first has to stand up and then walk, meaning that the agent easily falls down on the ground, which is difficult to predict. The episode length is 1000 steps starting from
randomized initial states. We use action repeat $R=4$ for the Cheetah-run task, and
$R=2$ for the Walker-walk task and the Quadruped-walk task.

\begin{figure*}[t]
  \centering
  \centerline{\includegraphics[width=0.75\textwidth]{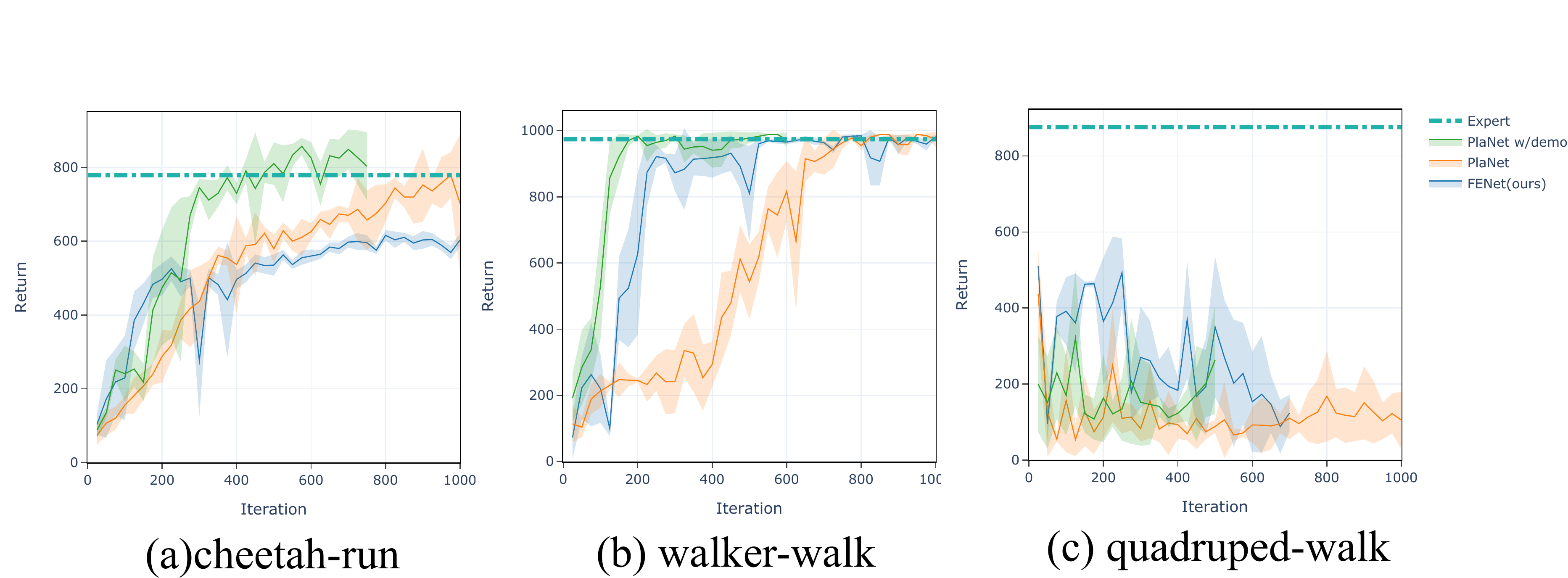}}
  \caption{Comparison of FENet to PlaNet and "PlaNet with demonstrations". Plots show test
  performance over learning iterations. The lines show means and the areas show standard deviations over 10 trajectories.}
  \label{exp1}
\end{figure*}

\subsection{Performance in standard visual control tasks}
We compare the performance of FENet to PlaNet \cite{hafner2018learning} and "PlaNet with demonstrations" in standard visual control tasks mentioned above. We use PlaNet as a baseline method because PlaNet is one of the most basic model-based RL methods using Recurrent State Space Model, on top of which we build our model. As FENet uses expert data, we create "PlaNet with demonstrations" for fair comparison. This variant of PlaNet has an additional experience replay pre-populated with expert trajectories and minimize a loss calculated from the expert data in addition to PlaNet's original loss.
Figure~\ref{exp1} shows that "PlaNet with demonstrations" is always better than PlaNet and that FENet is ranked higher as the difficulty of tasks gets higher. In Cheetah-run, FENet gives competitive performance with PlaNet. In Walker-walk, FENet and "PlaNet with demonstrations" are almost competitive, both of which are substantially better than PlaNet thanks to expert knowledge being leveraged to increase sample efficiency. In Quadruped-walk, FENet is slightly better than the other two baselines.

\subsection{Performance in sparse-reward visual control tasks}
\begin{figure*}[t]
  \centering
  \includegraphics[width=0.75\textwidth]{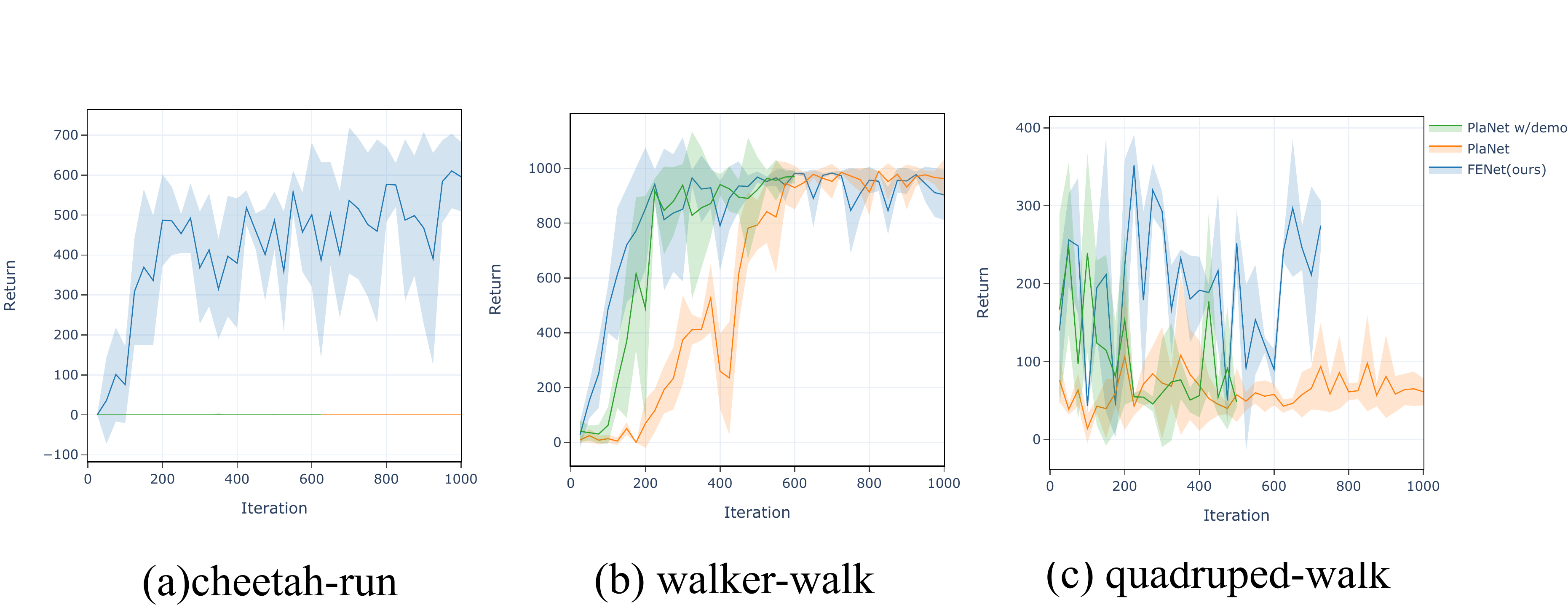}
  \caption{Comparison of FENet to PlaNet and "PlaNet with demonstrations" in sparse-reward settings, where agents do not get
  rewards less than 0.5. Plots show test performance over learning iterations.
  FENet substantially outperforms PlaNet. The lines show means and the areas show standard deviations over 10 trajectories.}
  \label{exp4}
\end{figure*}

In real-world robot learning, it is demanding to craft a dense reward function to lead robots to desired behaviors. It would
be helpful if an agent could acquire desired behaviors simply by giving sparse signals. We compare
the performance of FENet to PlaNet and "PlaNet with demonstrations" in sparse-reward settings, where agents do not get rewards less than 0.5 per time step
(Note that in the original implementation of Cheetah-run, Walker-walk and Quadruped-walk, agents get rewards ranging from 0 to 1 per time step).
Figure~\ref{exp4} shows that FENet outperforms PlaNet and "PlaNet with demonstrations" in all three tasks. In Cheetah-run, PlaNet and "PlaNet with demonstrations" are not able to get even a single reward.

\subsection{Performance with suboptimal experts}
\label{suboptimal_section}
\begin{figure}[thb]
  \centering
  \includegraphics[width=0.7\columnwidth]{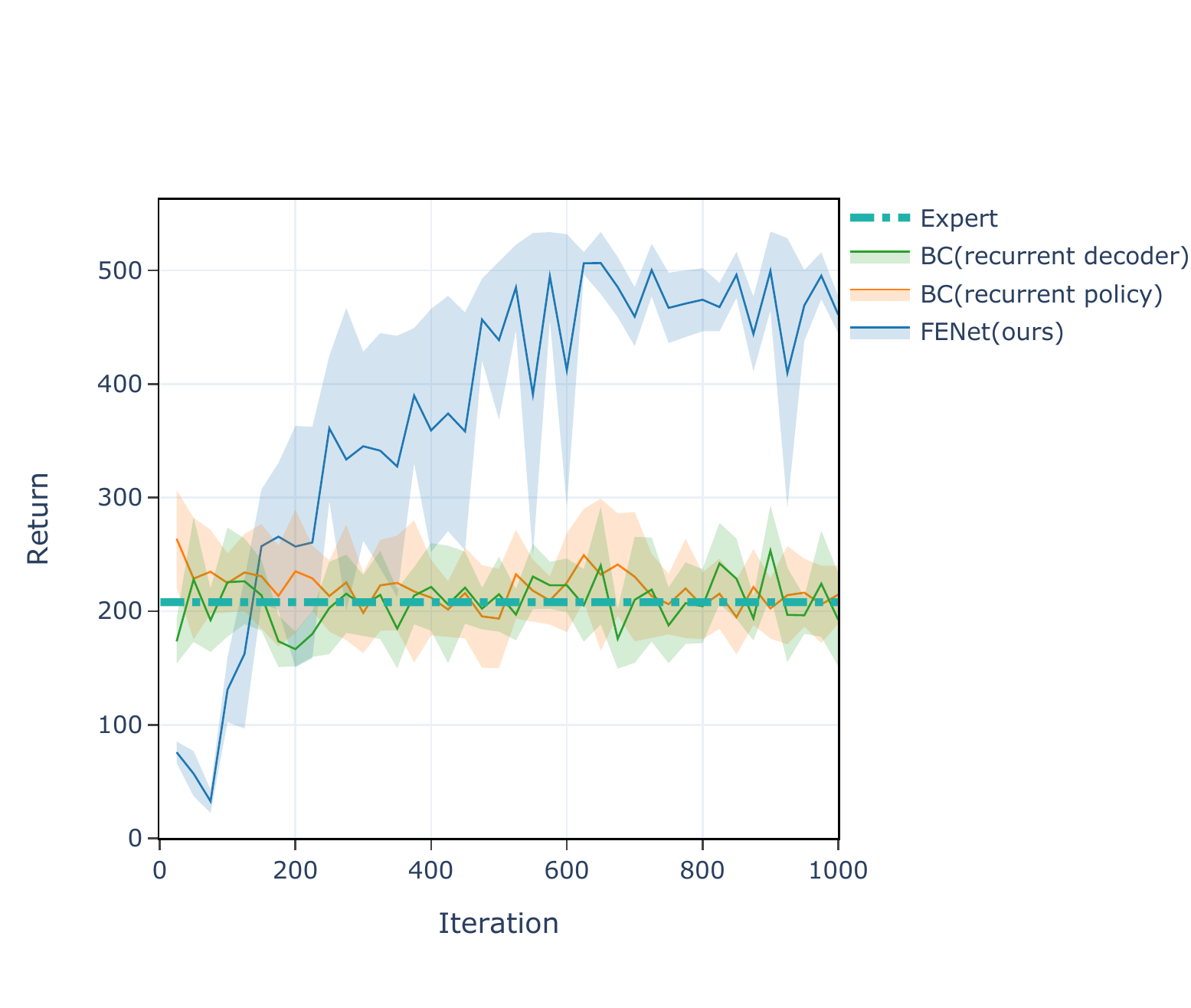}
  \caption{Comparison of FENet to IL methods when only suboptimal experts are available in Cheetah-run.
  Plots show test performance over learning iterations. Behavioral Cloning IL methods
  cannot surpass the suboptimal expert's return which FENet successfully surpasses. The lines show
  means and the areas show standard deviations over 10 trajectories.}
  \label{exp3}
  \vskip -0.2in
\end{figure}

In real-world robot learning, expert trajectories are often given by human experts. It is natural to assume that
expert trajectories are suboptimal and that there remains much room for improvement. We compare
the performance of FENet to Behavioral Cloning IL methods. We use two types of networks for behavioral cloning methods:
recurrent policy and recurrent decoder policy. The recurrent policy $\pi_R(a_t|o_t)$ is
neural networks with one gated recurrent unit cell and three dense layers. The
recurrent decoder policy $\pi_R(a_t,o_{t+1}|o_t)$ is neural networks with one gated recurrent unit cell and
four dense layers and deconvolution layers as in the decoder of PlaNet. Both networks does not get
raw pixel observations but take observations encoded by the same convolutional encoder as PlaNet's.
Figure~\ref{exp3} shows that while IL methods overfit to the expert and cannot surpass
the suboptimal expert performance, FENet is able to substantially surpass the suboptimal expert's performance.

\begin{algorithm}[H]
   \caption{Deep Free Energy Network (FENet)}
   \label{alg:FENet}
\begin{algorithmic}
   \STATE {\bfseries Input:}
   \STATE Seed episodes $S$ \text{\qquad \qquad \quad} Collect interval $C$
   \STATE Batch size $B$ \text{\qquad \qquad \qquad \ \ } Chunk length $L$
   \STATE Expert episodes $N$ \text{\qquad \qquad} Target smoothing rate $\rho$
   \STATE Learning rate $\alpha$
   \STATE State prior $p_\theta(u_t|h_t)$ \text{\quad \quad} State posterior $q_\phi(u_t|h_t, o_t)$
   \STATE Policy prior $p_\theta(a_t|u_t, h_t)$ \text{\ \ } Policy posterior $q_\psi(a_t|u_t, h_t)$
   \STATE Likelihood $p_\theta(o_t|u_t, h_t)$, $p_\theta(r_{t-1}|u_t, h_t)$
   \STATE Value function $V_\omega(u_t)$ \text{\ \ } Target value function $V_{\omega_{targ}}(u_t)$
   \STATE
   \STATE Initialize expert dataset $\mathcal{D}_E$ with $N$ expert trajectories
   \STATE Initialize agent dataset $\mathcal{D}_A$ with $S$ random episodes
   \STATE Initialize neural network parameters $\theta, \phi, \psi, \omega$ randomly
   \WHILE{not converged}
     \FOR{update step $c=1 .. C$}
       \STATE \texttt{// Imitation Learning (IL)}
       \STATE Draw expert data $\{(o_t, a_t, o_{t+1})_{t=k}^{k+L}\}_{i=1}^B \sim \mathcal{D}_E$
       \STATE Compute Free Energy $\mathcal{F}_{IL}$ from equation \ref{fil}
       \STATE \texttt{// Reinforcement Learning (RL)}
       \STATE Draw agent data $\{(o_t, a_t, r_t, o_{t+1})_{t=k}^{k+L}\}_{i=1}^B \sim \mathcal{D}_A$
       \STATE Compute Free Energy $\mathcal{F}_{RL}$ from equation \ref{frl}
       \STATE Compute $V$ function's Loss $\mathcal{L}$ from equation \ref{value_loss}
       \STATE \texttt{// Update parameters}
       \STATE $\theta \leftarrow \theta - \alpha \nabla_\theta (\mathcal{F}_{IL}+\mathcal{F}_{RL})$
       \STATE $\phi \leftarrow \phi - \alpha \nabla_\phi (\mathcal{F}_{IL}+\mathcal{F}_{RL})$
       \STATE $\psi \leftarrow \psi - \alpha \nabla_\psi (\mathcal{F}_{IL}+\mathcal{F}_{RL})$
       \STATE $\omega \leftarrow \omega - \alpha \nabla_\omega \mathcal{L}$
       \STATE $\omega_{targ} \leftarrow \rho \omega_{targ} + (1 - \rho) \omega$
     \ENDFOR
     \STATE \texttt{// Environment interaction}
     \STATE $o_1 \leftarrow$ \texttt{env.reset()}
     \FOR{time step $t=1 .. T$}
      \STATE Infer hidden states $u_t \leftarrow q_\phi(u_t|h_t, o_t)$
      \STATE Calculate actions $a_t \leftarrow q_\psi(a_t|u_t, h_t)$
      \STATE Add exploration noise to actions
      \STATE $r_t, o_{t+1} \leftarrow$ \texttt{env.step} $(a_t)$
     \ENDFOR
     \STATE $\mathcal{D}_A \leftarrow \mathcal{D}_A \cup \{(o_t, a_t, r_t, o_{t+1})_{t=1}^T\}$
   \ENDWHILE
\end{algorithmic}
\end{algorithm}

\begin{figure*}[t]
  \centering
  \centerline{\includegraphics[width=0.6\textwidth]{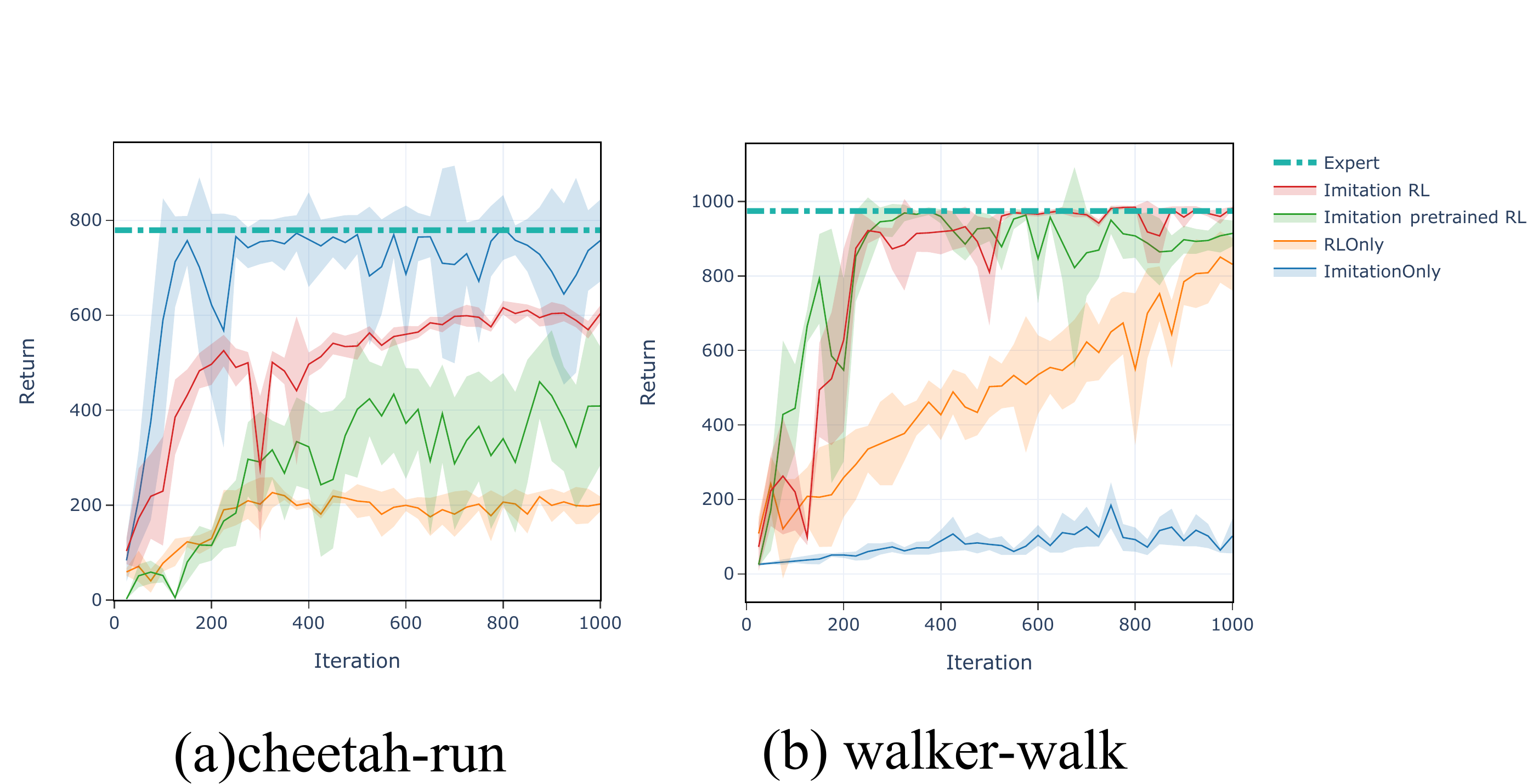}}
  \caption{Comparison of FENet (imitation RL) to partial FENet (as ablation studies: Imitation-pretrained RL, RL only, and Imitation only). Plots show test
  performance over learning iterations. The lines show means and the areas show standard deviations over 10 trajectories.}
  \label{exp2}
\end{figure*}

\subsection{Ablation Study}
Figure~\ref{exp2} compares FENet with other types of agents partially using FENet's loss in Cheetah-run and Walker-walk (ablation study). 'Imitation RL' is the proposed FENet agent that does
IL and RL at the same time, minimizing $\mathcal{F}_{IL}+\mathcal{F}_{RL}$. 'Imitation-pretrained RL' is an agent that first learns the model only with imitation (minimizing $\mathcal{F}_{IL}$)
and then does RL using the pre-trained model (minimizing $\mathcal{F}_{RL}$). 'RL only' is an agent that does RL only, minimizing $\mathcal{F}_{RL}$. 'Imitation only' is an agent that does IL only, minimizing $\mathcal{F}_{IL}$.
While 'Imitation only' gives the best performance and 'Imitation RL' gives the second best in Cheetah-run, 'Imitation RL' gives the best
performance and 'Imitation only' gives the worst performance in Walker-walk. We could say 'Imitation RL' is the most robust to the properties of tasks.


\section{Related Work}
\textbf{Active Inference} \quad
Friston, who first proposed Active Inference, has evaluated the performance in simple control tasks and a low-dimensional maze \cite{friston2012active, friston2015active}.
Ueltzhoffer implemented Active Inference with Deep Neural Networks and evaluated the performance in a simple control task \cite{ueltzhoffer2018deep}.
Millidge proposed a Deep Active Inference framework with value functions to estimate the correct Free Energy and succeeded in solving Gym environments \cite{millidge2019deep}.
Our approach extends Deep Active Inference to combine IL and RL, solving more challenging tasks.

\textbf{RL from demonstration} \quad
Reinforced Imitation Learning succeeds in reducing sample complexity by using imitation as pre-training before RL \cite{pfeiffer2018reinforced}.
Adding demonstrations into a replay buffer of off policy RL methods also demonstrates high sample efficiency \cite{vecerik2017leveraging, nair2018overcoming, paine2019making}.
Demo Augmented Policy Gradient mixes the policy gradient with a behavioral cloning gradient \cite{rajeswaran2017learning}.
Deep Q-learning from Demonstrations (DQfD) not only use demonstrations for pre-training but also calculates gradients from demonstrations and environment interaction data \cite{hester2018deep}.
Truncated HORizon Policy Search uses demonstrations to shape rewards so that subsequent planning can achieve superior performance to RL even when experts are suboptimal \cite{sun2018truncated}.
Soft Q Imitation Learning gives rewards that encourage the agent to return to demonstrated states in order to avoid policy collapse \cite{reddy2019sqil}.
Our approach is similar to DQfD in terms of mixing gradients calculated from demonstrations and from environment interaction data. One key difference is that
FENet concurrently learns the generative model of the world  so that it can be robust to wider environment properties.


\textbf{Control with latent dynamics model} \quad
World Models acquire latent spaces and dynamics over the spaces separately, and evolve simple linear controllers to solve visual control tasks \cite{ha2018world}.
PlaNet learns Recurrent State Space Model and does planning with Model Predictive Control at test phase \cite{hafner2018learning}.
Dreamer, which is recently built upon PlaNet, has a policy for latent imagination and achieved higher performance than PlaNet \cite{hafner2019dream}.
Our approach also uses Recurrent State Space Model to describe variational inference, and we are the first to unify IL and RL over latent
dynamics models to the best of our knowledge.

\section{Conclusion}
We present FENet, an agent that unifies Imitation Learning (IL) and Reinforcement Learning (RL)
using Free Energy objectives. For this, we theoretically extend the Free Energy Principle and introduce a policy prior that encodes experts' behaviors
and a policy posterior that learns to maximize expected rewards without deviating too much from the policy prior.
FENet outperforms model-based RL and RL with demonstrations especially in visual control tasks with sparse rewards and FENet also outperforms suboptimal experts' performance unlike Behavioral cloning. Strong potentials in sparse environment with suboptimal experts are important factors for real-world robot learning.

Directions for future work include learning the balance between IL and RL, i.e. Free Energy and Expected Free Energy so that
the agent can select the best approach to solve its confronting tasks by monitoring the value of Free Energy.
It is also important to evaluate FENet in real-world robotic tasks to fully manifest its effectiveness and reveal demands for further improvements.

\bibliography{example_paper}

\begin{thebibliography}{31}
\providecommand{\natexlab}[1]{#1}
\providecommand{\url}[1]{\texttt{#1}}
\expandafter\ifx\csname urlstyle\endcsname\relax
  \providecommand{\doi}[1]{doi: #1}\else
  \providecommand{\doi}{doi: \begingroup \urlstyle{rm}\Url}\fi

\bibitem[Friston(2010)]{friston2010free}
Friston, K.
\newblock The free-energy principle: a unified brain theory?
\newblock \emph{Nature reviews neuroscience}, 11\penalty0 (2):\penalty0
  127--138, 2010.

\bibitem[Friston et~al.(2006)Friston, Kilner, and Harrison]{friston2006free}
Friston, K., Kilner, J., and Harrison, L.
\newblock A free energy principle for the brain.
\newblock \emph{Journal of Physiology-Paris}, 100\penalty0 (1-3):\penalty0
  70--87, 2006.

\bibitem[Friston et~al.(2012)Friston, Samothrakis, and
  Montague]{friston2012active}
Friston, K., Samothrakis, S., and Montague, R.
\newblock Active inference and agency: optimal control without cost functions.
\newblock \emph{Biological cybernetics}, 106\penalty0 (8-9):\penalty0 523--541,
  2012.

\bibitem[Friston et~al.(2015)Friston, Rigoli, Ognibene, Mathys, Fitzgerald, and
  Pezzulo]{friston2015active}
Friston, K., Rigoli, F., Ognibene, D., Mathys, C., Fitzgerald, T., and Pezzulo,
  G.
\newblock Active inference and epistemic value.
\newblock \emph{Cognitive neuroscience}, 6\penalty0 (4):\penalty0 187--214,
  2015.

\bibitem[Ha \& Schmidhuber(2018)Ha and Schmidhuber]{ha2018world}
Ha, D. and Schmidhuber, J.
\newblock World models.
\newblock \emph{arXiv preprint arXiv:1803.10122}, 2018.

\bibitem[Haarnoja et~al.(2018)Haarnoja, Zhou, Abbeel, and
  Levine]{haarnoja2018soft}
Haarnoja, T., Zhou, A., Abbeel, P., and Levine, S.
\newblock Soft actor-critic: Off-policy maximum entropy deep reinforcement
  learning with a stochastic actor.
\newblock In \emph{International conference on machine learning}, pp.\
  1861--1870. PMLR, 2018.

\bibitem[Hafner et~al.(2019{\natexlab{a}})Hafner, Lillicrap, Ba, and
  Norouzi]{hafner2019dream}
Hafner, D., Lillicrap, T., Ba, J., and Norouzi, M.
\newblock Dream to control: Learning behaviors by latent imagination.
\newblock \emph{arXiv preprint arXiv:1912.01603}, 2019{\natexlab{a}}.

\bibitem[Hafner et~al.(2019{\natexlab{b}})Hafner, Lillicrap, Fischer, Villegas,
  Ha, Lee, and Davidson]{hafner2018learning}
Hafner, D., Lillicrap, T., Fischer, I., Villegas, R., Ha, D., Lee, H., and
  Davidson, J.
\newblock Learning latent dynamics for planning from pixels.
\newblock In Chaudhuri, K. and Salakhutdinov, R. (eds.), \emph{Proceedings of
  the 36th International Conference on Machine Learning}, volume~97, pp.\
  2555--2565, Long Beach, California, USA, 2019{\natexlab{b}}. PMLR.

\bibitem[Hester et~al.(2018)Hester, Vecerik, Pietquin, Lanctot, Schaul, Piot,
  Horgan, Quan, Sendonaris, Osband, et~al.]{hester2018deep}
Hester, T., Vecerik, M., Pietquin, O., Lanctot, M., Schaul, T., Piot, B.,
  Horgan, D., Quan, J., Sendonaris, A., Osband, I., et~al.
\newblock Deep q-learning from demonstrations.
\newblock In \emph{Thirty-Second AAAI Conference on Artificial Intelligence},
  2018.

\bibitem[Kapturowski et~al.(2019)Kapturowski, Ostrovski, Dabney, Quan, and
  Munos]{kapturowski2018recurrent}
Kapturowski, S., Ostrovski, G., Dabney, W., Quan, J., and Munos, R.
\newblock Recurrent experience replay in distributed reinforcement learning.
\newblock In \emph{International Conference on Learning Representations}, 2019.

\bibitem[Kingma \& Ba(2014)Kingma and Ba]{kingma2014adam}
Kingma, D.~P. and Ba, J.
\newblock Adam: A method for stochastic optimization.
\newblock \emph{arXiv preprint arXiv:1412.6980}, 2014.

\bibitem[Levine(2018)]{levine2018reinforcement}
Levine, S.
\newblock Reinforcement learning and control as probabilistic inference:
  Tutorial and review.
\newblock \emph{arXiv preprint arXiv:1805.00909}, 2018.

\bibitem[Millidge(2019)]{millidge2019deep}
Millidge, B.
\newblock Deep active inference as variational policy gradients.
\newblock \emph{arXiv preprint arXiv:1907.03876}, 2019.

\bibitem[Mnih et~al.(2013)Mnih, Kavukcuoglu, Silver, Graves, Antonoglou,
  Wierstra, and Riedmiller]{mnih2013playing}
Mnih, V., Kavukcuoglu, K., Silver, D., Graves, A., Antonoglou, I., Wierstra,
  D., and Riedmiller, M.
\newblock Playing atari with deep reinforcement learning.
\newblock \emph{arXiv preprint arXiv:1312.5602}, 2013.

\bibitem[Nair et~al.(2018)Nair, McGrew, Andrychowicz, Zaremba, and
  Abbeel]{nair2018overcoming}
Nair, A., McGrew, B., Andrychowicz, M., Zaremba, W., and Abbeel, P.
\newblock Overcoming exploration in reinforcement learning with demonstrations.
\newblock In \emph{2018 IEEE International Conference on Robotics and
  Automation (ICRA)}, pp.\  6292--6299. IEEE, 2018.

\bibitem[Nair \& Hinton(2010)Nair and Hinton]{nair2010rectified}
Nair, V. and Hinton, G.~E.
\newblock Rectified linear units improve restricted boltzmann machines.
\newblock In \emph{Proceedings of the 27th international conference on machine
  learning (ICML-10)}, pp.\  807--814, 2010.

\bibitem[Paine et~al.(2019)Paine, Gulcehre, Shahriari, Denil, Hoffman, Soyer,
  Tanburn, Kapturowski, Rabinowitz, Williams, et~al.]{paine2019making}
Paine, T.~L., Gulcehre, C., Shahriari, B., Denil, M., Hoffman, M., Soyer, H.,
  Tanburn, R., Kapturowski, S., Rabinowitz, N., Williams, D., et~al.
\newblock Making efficient use of demonstrations to solve hard exploration
  problems.
\newblock \emph{arXiv preprint arXiv:1909.01387}, 2019.

\bibitem[Paszke et~al.(2017)Paszke, Gross, Chintala, Chanan, Yang, DeVito, Lin,
  Desmaison, Antiga, and Lerer]{paszke2017automatic}
Paszke, A., Gross, S., Chintala, S., Chanan, G., Yang, E., DeVito, Z., Lin, Z.,
  Desmaison, A., Antiga, L., and Lerer, A.
\newblock Automatic differentiation in pytorch.
\newblock 2017.

\bibitem[Pfeiffer et~al.(2018)Pfeiffer, Shukla, Turchetta, Cadena, Krause,
  Siegwart, and Nieto]{pfeiffer2018reinforced}
Pfeiffer, M., Shukla, S., Turchetta, M., Cadena, C., Krause, A., Siegwart, R.,
  and Nieto, J.
\newblock Reinforced imitation: Sample efficient deep reinforcement learning
  for mapless navigation by leveraging prior demonstrations.
\newblock \emph{IEEE Robotics and Automation Letters}, 3\penalty0 (4):\penalty0
  4423--4430, 2018.

\bibitem[Pomerleau(1991)]{pomerleau1991efficient}
Pomerleau, D.~A.
\newblock Efficient training of artificial neural networks for autonomous
  navigation.
\newblock \emph{Neural computation}, 3\penalty0 (1):\penalty0 88--97, 1991.

\bibitem[Rajeswaran* et~al.(2018)Rajeswaran*, Kumar*, Gupta, Vezzani, Schulman,
  Todorov, and Levine]{rajeswaran2017learning}
Rajeswaran*, A., Kumar*, V., Gupta, A., Vezzani, G., Schulman, J., Todorov, E.,
  and Levine, S.
\newblock {Learning Complex Dexterous Manipulation with Deep Reinforcement
  Learning and Demonstrations}.
\newblock In \emph{Proceedings of Robotics: Science and Systems (RSS)}, 2018.

\bibitem[Reddy et~al.(2019)Reddy, Dragan, and Levine]{reddy2019sqil}
Reddy, S., Dragan, A.~D., and Levine, S.
\newblock Sqil: imitation learning via regularized behavioral cloning.
\newblock \emph{arXiv preprint arXiv:1905.11108}, 2019.

\bibitem[Rhinehart et~al.(2018)Rhinehart, McAllister, and
  Levine]{rhinehart2018deep}
Rhinehart, N., McAllister, R., and Levine, S.
\newblock Deep imitative models for flexible inference, planning, and control.
\newblock \emph{arXiv preprint arXiv:1810.06544}, 2018.

\bibitem[Silver et~al.(2016)Silver, Huang, Maddison, Guez, Sifre, Van
  Den~Driessche, Schrittwieser, Antonoglou, Panneershelvam, Lanctot,
  et~al.]{silver2016mastering}
Silver, D., Huang, A., Maddison, C.~J., Guez, A., Sifre, L., Van Den~Driessche,
  G., Schrittwieser, J., Antonoglou, I., Panneershelvam, V., Lanctot, M.,
  et~al.
\newblock Mastering the game of go with deep neural networks and tree search.
\newblock \emph{nature}, 529\penalty0 (7587):\penalty0 484--489, 2016.

\bibitem[Sun et~al.(2018{\natexlab{a}})Sun, Bagnell, and
  Boots]{sun2018truncated}
Sun, W., Bagnell, J.~A., and Boots, B.
\newblock Truncated horizon policy search: Combining reinforcement learning \&
  imitation learning.
\newblock In \emph{International Conference on Learning Representations},
  2018{\natexlab{a}}.

\bibitem[Sun et~al.(2018{\natexlab{b}})Sun, Gordon, Boots, and
  Bagnell]{sun2018dual}
Sun, W., Gordon, G.~J., Boots, B., and Bagnell, J.
\newblock Dual policy iteration.
\newblock In \emph{Advances in Neural Information Processing Systems}, pp.\
  7059--7069, 2018{\natexlab{b}}.

\bibitem[Sutton et~al.(1998)Sutton, Barto, et~al.]{sutton1998introduction}
Sutton, R.~S., Barto, A.~G., et~al.
\newblock \emph{Introduction to reinforcement learning}, volume 135.
\newblock MIT press Cambridge, 1998.

\bibitem[Tassa et~al.(2018)Tassa, Doron, Muldal, Erez, Li, Casas, Budden,
  Abdolmaleki, Merel, Lefrancq, et~al.]{tassa2018deepmind}
Tassa, Y., Doron, Y., Muldal, A., Erez, T., Li, Y., Casas, D. d.~L., Budden,
  D., Abdolmaleki, A., Merel, J., Lefrancq, A., et~al.
\newblock Deepmind control suite.
\newblock \emph{arXiv preprint arXiv:1801.00690}, 2018.

\bibitem[Ueltzh{\"o}ffer(2018)]{ueltzhoffer2018deep}
Ueltzh{\"o}ffer, K.
\newblock Deep active inference.
\newblock \emph{Biological cybernetics}, 112\penalty0 (6):\penalty0 547--573,
  2018.

\bibitem[Vecerik et~al.(2017)Vecerik, Hester, Scholz, Wang, Pietquin, Piot,
  Heess, Roth{\"o}rl, Lampe, and Riedmiller]{vecerik2017leveraging}
Vecerik, M., Hester, T., Scholz, J., Wang, F., Pietquin, O., Piot, B., Heess,
  N., Roth{\"o}rl, T., Lampe, T., and Riedmiller, M.
\newblock Leveraging demonstrations for deep reinforcement learning on robotics
  problems with sparse rewards.
\newblock \emph{arXiv preprint arXiv:1707.08817}, 2017.

\bibitem[Verma et~al.(2019)Verma, Le, Yue, and Chaudhuri]{verma2019imitation}
Verma, A., Le, H., Yue, Y., and Chaudhuri, S.
\newblock Imitation-projected programmatic reinforcement learning.
\newblock In \emph{Advances in Neural Information Processing Systems}, pp.\
  15726--15737, 2019.

\end{thebibliography}
\bibliographystyle{icml2021}

\appendix
\section{Implementation}
To stabilize the learning process, we adopt burn-in, a technique to recover initial states of RNN's
hidden variables $h_t$ \citep{kapturowski2018recurrent}. As shown in Algorithm 1,
the agent calculates the Free Energy with mini batches sampled from the expert or agent
dataset $\mathcal{D}$, which means that $h_t$ is initialized randomly in every mini batch calculation.
Since the Free Energy heavily depends on the value of $h_t$, it is crucial to estimate the accurate hidden states every iteration.
We set a burn-in period when a portion of the mini batch data sequence is used for unrolling the networks to produce initial states of $h_t$.
After the burn-in period, we update the networks using the remaining part of the data sequence.

We use PyTorch \citep{paszke2017automatic} to write neural networks and run experiments using NVIDIA GeForce GTX 1080 Ti / RTX 2080 Ti / Tesla V100 GPU (1 GPU per experiment).
The training time for our FENet implementation is about 24 hours on the DeepMind Control Suite environment.
As for the hyper parameters, we use the convolutional encoder and decoder networks from \citep{ha2018world} and
Recurrent State Space Model from \citep{hafner2018learning} and implement all other functions as
three dense layers of size 200 with ReLU activations \citep{nair2010rectified}. We made a design choice to make the
policy prior, the policy posterior, the observation likelihood, and the reward likelihood deterministic functions
while we make the state prior and the state posterior stochastic functions. We use the batch size $B=25$
for 'Imitation RL' with FENet, and $B=50$ for other types and baseline methods.
We use the chunk length $L=50$, the burn-in period 20. We use seed episodes $S=40$,
expert episodes $N=10000$ trained with PlaNet \citep{hafner2018learning}, collect interval $C=100$ and action exploration noise Normal(0, 0.3).
We use the discount factor $\gamma = 0.99$ and the target smoothing rate $\rho = 0.01$.
We use Adam \citep{kingma2014adam} with learning rates $\alpha = 10^{-3}$ and scale down gradient norms that exceed 1000.
We scale the reward-related loss by 100, the policy-prior-related loss by 10. We clip KL loss between
the hidden states below 3 free nats and clip KL loss between the policies below 0.6.

\section{Expert data collection process}
We first trained PlaNet for Cheetah-run and Walker-walk. We used Soft Actor-Critic (SAC) \citep{haarnoja2018soft} for Quadruped-walk because PlaNet cannot solve Quadruped-walk very much as shown in Figure~\ref{exp1}. Then we saved the model parameters when PlaNet or SAC achieved asymptotic performance for each task. After this, we generated 10,000 expert trajectories for each task using the saved model parameters. The suboptimal expert dataset is what we collected by using the model parameters of fewer learning iterations before reaching to asymptotic performance. For example as shown in Figure~\ref{exp3}, the suboptimal expert in Cheetah-run is PlaNet agent that was trained halfway to reach the return of 200.

\end{document}